\documentclass[runningheads]{llncs}
\usepackage{graphicx}
\usepackage{amsmath}
\usepackage{amscd}
\usepackage{amssymb}
\usepackage{color}
\usepackage{appendix}

\newcommand{\R}{\mathbb{R}}

\newcommand{\bE}{\mathbb{E}}
\newcommand{\cb}{{\mathcal{B}}}

\newcommand{\al}{\alpha}
\newcommand{\be}{\beta}

\newcommand{\de}{\delta}
\newcommand{\ep}{\mathcal{E}}

\newcommand{\id}{\mathrm{id}}

\newcommand{\rk}{\mathrm{rk}}
\newcommand{\fhat}{\widehat{f}}

\newcommand{\beq}{\begin{equation}}
\newcommand{\eeq}{\end{equation}}
\newcommand{\lra}{\longrightarrow}

\begin{document}

\title{A geometric interpretation of stochastic gradient descent using diffusion metrics}
%\thanks{Supported by Amazon and MSCA funded project GHAIA (GA 777822).}

\author{R. Fioresi\inst{1} \and
P. Chaudhari\inst{2} \and
S. Soatto\inst{3}}
\authorrunning{Fioresi et al.}
\institute{
Dipartimento di Matematica, piazza Porta
San Donato 5, University of Bologna, 40126 Bologna, Italy\\
\email{rita.fioresi@unibo.it}
\and
Department of Electrical and Systems Engineering,
University of Pennsylvania  
\email{pratikac@seas.upenn.edu}
\\
\and
Computer Science Department,
University of California, Los Angeles
\email{soatto@cs.ucla.edu}}
\maketitle              % typeset the header of the contribution
\begin{abstract}
Stochastic  gradient  descent  (SGD)  is  a key ingredient
in the training of deep neural networks and yet its geometrical
significance appears elusive.
%We model SGD descent using a 
We study a deterministic
model in which the trajectories of our
dynamical systems are described via geodesics 
of a family of metrics arising from the diffusion matrix. 
These metrics encode information about 
the highly non-isotropic
gradient noise in SGD. We establish a parallel with General
Relativity models, where the role of the
electromagnetic field is played by the gradient of the loss function.
We compute an example of 
a two layer network.
%to provide empirical validation of 
%our results.

\end{abstract}

\section{Introduction}

In this paper, we interpret the
diffusion matrix for SGD (see \cite{cs}),
as a metric for the parameters' space; thus we
provide a deterministic equation (\ref{sgd-gr}) %approximating effectively, 
that we compare, near the equilibrium points, to the stochastic
gradient descent (\ref{sgd-eq}). We start with the definition of the diffusion
matrix $D(x)$ and then we write it in the form (\ref{diff-expr1}),
that shows clearly how $D(x)$ captures 
in an essential way the anisotropy of the dynamical system
ruling the SGD. In other words, in which way $D(x)$ 
is one of key factors in the difference between the
 steady state solutions of SGD from those of %domain determined by
ordinary GD (see comparison in \cite{cs}). %From the explicit form for 
Using the diffusion matrix, we then
define a family of metrics on the parameters' space, that
we call {\sl diffusion metrics}.  
We then take euristically
Einstein equation describing
the geodesic on a Riemannian manifold, 
%with metric defined via the diffusion matrix, 
for the motion of a particle
subject to gravity and electromagnetic field and we replace the electromagnetic
field contribution by the forcing term represented by 
the ordinary gradient, the gravity taken into account
by the diffusion metric itself. After some mild hypotheses on the
network, we obtain that the geodesics with respect
to this equation, correspond precisely to the evolution of
a dynamical system, which not subject to euclidean gradient descent, but to
relativistic gradient descent (RGD) with respect to the family of diffusion metrics.

In the end we obtain the equation \ref{sgd-gr}, which is along
the same vein as in \cite{amari}, that is, natural gradient descent,
but whose significance is much deeper in the context of SGD, since it stems from the
anisotropy of the loss with respect to the various parameters, which
%responsible for 
encodes the difference in the dynamics between GD and SGD.
We also compare our result with the ones in \cite{cs} and show
they are perfectly compatible.
%We finally illustrate our theory with some examples. 

\section{Continuous-time SGD and Diffusion matrix}

Stochastic gradient descent performs un update of the weights 
$x$  %\in \Omega \subset \R^d$
of a neural network, replacing the ordinary gradient of the
loss function $f=\sum_{i=1}^N f_i$ with $ \nabla_\cb f$:
\beq\label{sgd-eq}
dx=-\nabla_\cb f dt, \qquad 
\nabla_\cb f=\frac{1}{|\cb|} \sum_{i \in \cb} \nabla f_i
\eeq
where $dx$ represents the continuous version of
the weight update at step $j$: $x_{j+1}=x_j-\eta\nabla_\cb f(x_j)$,
with the learning rate $\eta$ incorporated into the expression of
$\nabla_\cb f$ and $\cb$ is the mini-batch.
%\comment{R please check if right or if $\eta=1$}
In the expression of the loss function
$f=\sum_{i=1}^N f_i$, $f_i$ is the loss relative
to the $i$-th element in our dataset $\Sigma$ of size $|\Sigma|=N$.

We assume that
weights belong to a compact subset $\Omega \subset \R^d$
and that the $f_i$'s satisfy suitable regularity conditions
(see \cite{cs} Sec. 2 for more details).

We define the \textit{diffusion matrix} as the product of 
the size of the mini-batch $|\cb|$ and the
variance of $\nabla_\cb f$, viewed as a random variable,
$\phi: \Sigma \lra \R^d$,  $\phi(z_i)=\nabla f_i$:
\beq \label{diff-def}
D(x)=\bE[(\phi-\bE[\phi])(\phi-\bE[\phi])^t]
\eeq
Notice that $D(x)$ is $d \times d$ matrix
independent from the mini-batch size; it 
only depends on the weights $x$ and loss function $f$ and the dataset 
$\Sigma$. 
With a direct calculation one shows that:%\comment{R. we may write more}
\beq\label{diff-expr1}
\begin{array}{rl}
D&=\frac{1}{N}\sum_k (\nabla f_k) (\nabla f_k)^t-(\nabla f) (\nabla f)^t 
=\frac{1}{N^2}(\langle \partial_r \fhat, \partial_s \fhat \rangle)
\end{array}
\eeq
where: 
$$
\fhat=(f_1 -  f_2, f_1-f_3, \dots, f_{N-1} -   f_N)\in \R^{N(N-1)/2}
$$
and $\langle,\rangle$ is the euclidean scalar product. In fact:
$$
\begin{array}{rl}
D_{rs}&=\frac{1}{N}\sum_{k=1}^N \partial_rf_k \,  \partial_s f_k-
\frac{1}{N^2}\sum_{i,j=1}^N \partial_rf_i \,  \partial_s f_j\\ \\
&=\frac{1}{N^2}[N( \partial_rf_1 \,  \partial_s f_1+\dots+
 \partial_rf_N \,  \partial_s f_N)+\\ \\
&-(\partial_rf_1 \,  \partial_s f_1+
\partial_rf_1 \,  \partial_s f_2+\dots+\partial_rf_N \,  \partial_s f_N)]= 
\\ \\
&=\frac{1}{N^2}[(N-1)\partial_rf_1 \,  \partial_s f_1-
\partial_rf_1 \,  \partial_s f_2-\dots -\partial_rf_1 \,  \partial_s f_N+\\ \\
&-\partial_rf_2 \,  \partial_s f_1+(N-1)\partial_rf_2 \,  \partial_s f_2-
\dots -\partial_rf_2 \,  \partial_s f_N+ \dots \\ \\
&-\partial_rf_N \,  \partial_s f_1- \partial_rf_N \,  \partial_s f_2+
\dots +(N-1)\partial_rf_N \,  \partial_s f_N]
\end{array}
$$
which gives (almost immediately):
$$
\begin{array}{rl}
D_{rs}&=\frac{1}{N^2}[
(\partial_rf_1 -  \partial_r f_2)(\partial_sf_1 -\partial_s f_2)+
(\partial_rf_1 -  \partial_r f_3)(\partial_sf_1 -\partial_s f_3)+\dots \\ \\
&+(\partial_rf_1 -  \partial_r f_N)(\partial_sf_1 -\partial_s f_N)+
(\partial_rf_2 -  \partial_r f_3)(\partial_sf_2 -\partial_s f_3)+
\dots \\ \\
&+(\partial_rf_{N-1} -  \partial_r f_N)(\partial_sf_{N-1} -\partial_s f_N)=\frac{1}{N^2}(\langle \partial_r \fhat, \partial_s \fhat \rangle)
\end{array}
$$
The diffusion matrix measures effectively the {\sl anisotropy}
of our data: $D=0$ if and only if $\partial_r(f_i)=\partial_r(f_j)$
for all $r=1, \dots, d$ and $i,j =1, \dots, N$. In other words
%\comment{R. personal interpretation, change if you do not agree}
the diffusion matrix measures how the loss of each datum
depends, at first order, on the weights in different way 
with respect to the loss of each of the other data. So, it tells us how much
we should expect the SGD dynamics to differ from the GD one.

Notice that the expression (\ref{diff-expr1}) 
gives us immediately 
%\comment{R. may write more} 
a bound on the
rank of $D$, namely $\rk(D)\leq N-1$. 
\begin{table}
\caption{Values for $N$ and $d$ for
various architectures on CIFAR and SVHN datasets (see \cite{hlmw}).}\label{tab1}
\begin{tabular}{|l|l|l|l|l|}
\hline
Architecture &  $d=|$Weights$|$ & $N=|$Data$|$, CIFAR & $N=|$Data$|$, SVHN \\ 
%$N$ ImageNet\\
\hline
ResNet &  1.7M & 60K & 600K \\
Wide ResNet &  11M & 60K & 600K\\
DenseNet (k=12)  & 1M &  60K & 600K \\
DenseNet (k=24) & 27.2M  & 60K & 600K \\ %& 1.2M\\
\hline
\end{tabular}
\end{table}

This table suggests that in many algorithms currently available,
the diffusion matrix has low rank, hence it is singular; this just
by comparing the size $d$ of $D$ and its rank which
bound by $N$. This fact turns out to be very important
in the construction of the {\sl diffusion metrics}, that we
will see below.
%We shall see in the
%next section that this is even more the case for the special
%case of a two layer network.

%\section{Diffusion matrix for a $2$-layer network}
%\comment{R. not sure we want to include it, file dm-expr2.tex}

\section{Diffusion metrics and General Relativity}
The evolution of a dynamical system
in general relativity takes place along the geodesics
according to the metric imposed on the Minkowski space by
the presence of gravitational masses. % and electromagnetic fields.
The equation for such geodesics, once Einstein equation is solved, is:
\beq\label{geod}
\frac{d^2x^\mu}{dt^2}\, + \, \Gamma_{\rho\sigma}^\mu
\frac{dx^\rho}{dt}
\frac{dx^\sigma}{dt}=\frac{q}{m}F^\mu_\nu \frac{dx^\nu}{dt}
\eeq
where $\Gamma_{\rho\sigma}^\mu$ are the Christoffel symbols for the
Levi-Civita connection:
\beq\label{lc-conn}
\Gamma_{uv}^{w}=
{\frac {1}{2}}g^{wz}
\left(\partial_u g_{vz}-\partial_z  g_{uv}+\partial_v g_{uz}\right) 
\eeq
and $\frac{q}{m}F^\mu_\nu$ is a term regarding an external
force, e.g. one coming from an electromagnetic
field.

\medskip
If we take time derivative of the differential equation ruling the ordinary 
(i.e. non stochastic) gradient descent:
$$
\frac{d^2x^\mu}{dt^2}\, = -\frac{d}{dt}\, \partial_\mu f
$$
and we compare with (\ref{geod}),
it is clear that  $-\frac{d}{dt}\, \partial_\mu f$
effectively replaces the force term $(q/m)F^\mu_\nu\frac{dx^\nu}{dt}$.
Hence, the geodesic equation (\ref{geod}) models 
the ordinary GD equation, if we take
a constant metric and we replace the force term with
the gradient of the loss; furthermore this
corresponds to the condition $D=0$ in SGD dynamics  
equation (\ref{sgd-eq}).

This suggests
heuristically to define a metric, 
depending on the diffusion matrix, which becomes constant
when $D=0$. As a side remark, notice that since $D$ is 
singular in many important pratical
applications (see Table \ref{tab1}), it is not reasonable
to use it to define the metric itself. 
On the other hand, 
%\comment{R. heuristic personal explanation, may remove}
since $D$ measures
the anisotropy of the weight space, it is reasonable to employ 
it to perturb the euclidean metric. So the stochastic nature
of the dynamical system ruled by the SGD
is replaced by a perturbation of the dynamics for the ordinary gradient
descent.  As an analogy, the
presence of (small) masses in space, 
in the weak field approximation (see \cite{ab}) of general relativity, 
generate gravity, hence a (small) deformation of
the euclidean metric.

\medskip
At each point $x\in \Omega$, we define a metric
called \textit{diffusion metric} as
\beq\label{dm-def}
g(x)=\id + \ep(x) D(x)
\eeq
%where $\ep$ is a real parameter, 
with $\ep(x)< 1/M_x$, where $M_x=\mathrm{max} \{\lambda\}$ 
with $\lambda$ eigenvalues of $D(x)$. This ensures that 
$g$ is non singular at each point. %Heuristically, we may
%think of $\ep$ as a real parameter %$\ep << 1$, so that
So we are defining a family of metrics 
depending on the real parameter $\ep$.

We expect this model to approximate the Fokker Planck solution
when $\be^{-1}$ is very small (see \cite{cs} for the notation
and more details).
%We plan to explore this correspondence more in detail in
%a forthcoming paper.
%\comment{R. I think $\ep$ may be thought as $\be$}

\medskip
Notice that our heuristic hypothesis on $\ep$ allows us to make the
so called {\sl weak field approximation} (see \cite{ab}):
$$
g^{-1}= \id - \ep D(x)
$$
Hence, we have
the following expression for
the Christoffel's symbols (in this
approximation we discard ${\mathcal{O}}(\ep^2)$):
\beq\label{c-symb}
\Gamma_{uv}^{w}= 
{\frac{1}{2}}\sum_z(\de_{wz} - \ep d_{wz}) \, \ep \,
\left(\partial_u d_{vz}-\partial_z  d_{uv}+\partial_v d_{uz}\right) =
{\frac {\ep}{2}}
\left(\partial_u d_{vw}-\partial_w  d_{uv}+\partial_v d_{uw}\right)
\eeq
where $d_{ij}$ are the coefficients of $D$,
and $\de_{wz}$ is the Kronecker delta.

\medskip
Let us now compute the Christoffel's symbols and then substitute them
into the geodesic equation (\ref{geod}).
$$
\begin{array}{rl}
\Gamma_{ij}^{k}&=
 \frac{\ep}{2N^2}
[\langle \partial_i \partial_j\fhat,\partial_k \fhat  \rangle+
\langle \partial_j \fhat,\partial_i \partial_k \fhat  \rangle-
\langle \partial_i \partial_k\fhat,\partial_j \fhat  \rangle-
\langle \partial_i \fhat,\partial_k \partial_j \fhat  \rangle+\\ \\
&+\langle \partial_j \partial_k\fhat,\partial_i \fhat  \rangle+
\langle \partial_k \fhat,\partial_i \partial_j \fhat  \rangle]=
 \frac{\ep}{N^2}\langle \partial_i \partial_j\fhat,\partial_k \fhat  \rangle.
\end{array}
$$
Let us substitute in (\ref{geod}) (writing the sum now):
\beq\label{geod2}
\frac{d^2x^k}{dt^2}\, + \,\frac{\ep}{N^2}\sum_{i,j}
\langle \partial_i \partial_j\fhat,\partial_k \fhat  \rangle
\frac{dx^i}{dt}
\frac{dx^j}{dt}=\frac{d}{dt}\partial_k f
\eeq
Let us concentrate on the expression:
$$
\begin{array}{rl}
\frac{\ep}{N^2}\sum_{i,j}
\langle \partial_i \partial_j\fhat,\partial_k \fhat  \rangle
\frac{dx^i}{dt}
\frac{dx^j}{dt}&=
\frac{\ep}{N^2}\sum_{i,j,\al}
\partial_i \partial_j\fhat_\al \, \partial_k \fhat_\al
\frac{dx^i}{dt}
\frac{dx^j}{dt}= \\ \\
&=\frac{\ep}{N^2}\sum_{\al}
\frac{d^2\fhat_\al}{dt^2}\partial_k \fhat_\al
\end{array}
$$
Now we take the integral in $dt$ (we compute by parts twice):
$$
\begin{array}{rl}
 \frac{\ep}{N^2}\int\sum_{\al}
\frac{d^2\fhat_\al}{dt^2}\partial_k \fhat_\al \, dt&=
 \frac{\ep}{N^2}\sum_{\al}\left[\frac{d\fhat_\al}{dt}\partial_k \fhat_\al  -
\int \frac{d\fhat_\al}{dt}\frac{\partial_k \fhat_\al}{dt} \, dt\right]= \\ \\
&=\frac{\ep}{N^2}\sum_{\al}\left[\frac{d\fhat_\al}{dt}\partial_k \fhat_\al  -
\fhat_\al\frac{d}{dt}\partial_k \fhat_\al+
\int \fhat_\al\frac{d^2}{dt^2}\partial_k \fhat_\al \, dt\right]
\end{array}
$$
Notice that, in many practical applications we have:
\beq\label{approx1}
\frac{d^2}{dt^2}\partial_k \fhat_\al=0
\eeq
because $\partial_i\partial_j\partial_k \fhat=0$. 
%\bigskip\hrule\medskip
%RITA: in the 2 layer network example, since each loss
%depends quadratically from the weights, we clearly
%have (\ref{approx1}). We would need the sec. on 2 layer
%networks to fully justify this. For the proceeding
%it may not be worth it.
%\medskip\hrule\bigskip

Hence
$$
\begin{array}{rl}
 \frac{\ep}{N^2}\int\sum_{\al}
\frac{d^2\fhat_\al}{dt^2}\partial_k \fhat_\al \, dt&=
\frac{\ep}{N^2}\sum_{\al,\ell}
\left[\partial_\ell\fhat_\al \frac{dx^\ell}{dt} \partial_k \fhat_\al  -
\fhat_\al\frac{d}{dt}\partial_k \fhat_\al\right]=\\ \\
&=\ep \sum_{\ell} d_{k,\ell}\frac{dx^\ell}{dt}-
\frac{\ep}{N^2}\sum_{\al}\fhat_\al\frac{d}{dt}\partial_k \fhat_\al
\end{array}
$$
We now substitute the obtained expression into the
eq. (\ref{geod2}), taking the integral in $dt$:
$$
\frac{dx^k}{dt}\, + \ep \sum_{\ell} d_{k,\ell}\frac{dx^\ell}{dt}
-\frac{\ep}{N^2}\sum_{\al}\fhat_\al\frac{d}{dt}\partial_k \fhat_\al
=-\partial_k f
$$
We may assume $\frac{\ep}{N^2}$ very small, as motivated by
Table \ref{tab1}, hence discard this term.
Writing the equation into vector form, we have:
$$
\frac{dx}{dt}+\ep D\frac{dx}{dt}=-\nabla f
$$
Since by the weak field approximation 
$(I+\ep D)^{-1}\cong (I-\ep D)$,  we can write:
\beq\label{sgd-gr}
\frac{dx}{dt}=-(I-\ep D)\nabla f=-\nabla_D f
\eeq
where $\nabla_D f$ is the gradient computed according to the
diffusion metric (\ref{dm-def}).

We can summarize our result as follows: \textit{the SGD equation
(\ref{sgd-eq}) can be replaced, provided the approximations
(\ref{approx1}) holds, by the deterministic
equation (\ref{sgd-gr}), where the dynamical system evolves
with respect to the gradient computed according to the diffusion metric
(\ref{dm-def})}.

\medskip
%Hence we have produced an equation that we may use in place
%of the SGD dynamics (\ref{sgd-eq}). 
We now want to compare
our result (\ref{sgd-gr}) with \cite{cs} Sec. 1, in order to understand
how the steady state solutions of (\ref{sgd-gr}) compare to the SGD
steady state solutions described by (\ref{sgd-eq}).

In \cite{cs} the authors regard SGD as minimizing the function $\Phi$
instead of our loss $f$. Let us focus on eq. (8) in 
\cite{cs}, where the relation between $f$ and $\Phi$ is discussed.
In our case, we take  $\nabla \Phi= \nabla_D f$ so that eq. (\ref{sgd-gr})
becomes
\beq\label{compar-cs}
-\nabla f(x) + \widetilde{D}(x) \nabla \Phi(x) =0
\eeq
where $\widetilde{D}(x)=(I+\ep D(x))$ is the diffusion metric.

If the term $D(x)$ in (8) in \cite{cs}
is spelled out as our $\widetilde{D}(x)$, we
can write such equation as:
\beq\label{compar-cs2}
j(x)=-\nabla f(x) + \widetilde{D}(x) \nabla \Phi(x)  -\be^{-1} \nabla 
\cdot \widetilde{D}(x) 
\eeq
Notice that according to our approximation (\ref{approx1}), $\nabla 
\cdot \widetilde{D}(x)=0$. Hence, eq. (\ref{compar-cs2}) (that is
(8) in \cite{cs}) is
perfectly compatible with our treatment and furthermore the assumption
4 in \cite{cs} is fully justified by the fact $j(x)=0$.

% we expect 
%Hence, this system will
%evolve at $t \to \infty$ towards 
%$\rho^{\mathrm ss}$ the steady state distribution
%of the Fokker-Planck equation as detailed
%in \cite{cs}. %\comment{R. this should be tested, see next sec.}

% \section{Examples}

% In this section we consider a twolayer
% fully-connected network on MNIST (LeCun et al., \cite{lbbh}) in order to
% analyze trajectories of SGD and compare them with
% trajectories of (\ref{sgd-gr}) near equilibrium points.

\section{Conclusions}

The General Relativity model helps to provide with a
deterministic approach to the evolution of the dynamical system
described by SGD. 
%and to predict the most likely trajectories
%and the steady state solutions. 
The results are compatible with
\cite{cs}.

% \bigskip
% {\bf Acknowledgments}. We wish to thank Dr. A. Achille and 
% Prof. F. Faglioni for many illuminating discussions. R.F. wishes
% also to thank UCLA and Amazon for the wonderful hospitality
% while this work was done.

%\comment{R. please add entries to biblio}

%\usepackage{appendix}
\appendix
\section{Riemannian Geometry}

We collect few well known facts of Riemannian geometry,
inviting the reader to consult \cite{petersen} for more details.

\medskip
In Riemannian geometry, we define a metric $g$ on a smooth
manifold $M$, as a smooth assignment $p \mapsto g_p$, which
gives, for each $p \in M$, a (non degenerate)
scalar product on $T_p(M)$, the tangent
space of $M$ at $p$. Usually, this scalar product is assumed to be
positive definite,
however, for general relativity, it is necessary to drop this assumption,
so that we speak of a {\sl pseudometric}, because our main
example is the Minkowski metric. To ease the terminology, we say
``metric'', to include also this more general setting.

Once a metric is given, we say that $M$ is a {\sl Riemannian manifold}.
In local coordinates, $x^1,\dots, x^n$, we express the metric using $1$-forms:
$$
{\displaystyle g=\sum _{i,j}g_{ij}\,\mathrm {d} x^{i}\otimes \mathrm {d} x^{j}.}
$$
where
$$
\displaystyle g_{ij}|_{p}:=
  g_{p}\left(\left.{\frac {\partial }{\partial x^{i}}}\right|_{p},
  \left.{\frac {\partial }{\partial x^{j}}}\right|_{p}\right) \qquad
\hbox{and} \qquad 
\left\{ 
\frac{\partial }{\partial x^{1}}|_{p},\dots ,
\frac{\partial }{\partial x^{n}}|_{p}
\right\}
$$

is a basis of the tangent space $T_pM$.

For example $\R^n$, identified with its tangent space at every point,
has a canonical or standard metric given by:
$$
{\displaystyle g_{p}^{\mathrm {can} }\colon 
T_{p}\R^n\times T_{p}\R^n\longrightarrow \mathbf {R} ,
\qquad \left(\sum _{i}a_{i}{\frac {\partial }
{\partial x^{i}}},\sum _{j}b_{j}{\frac {\partial }
{\partial x^{j}}}\right)\longmapsto \sum _{i}a_{i}b_{i}.} 
$$
Here, $g_{ij}^{\mathrm {can} }=\delta _{ij}$.

Usually, we drop the $\sum$ symbol, following Einstein convention.

An {\sl affine connection} $\nabla$ on a smooth manifold $M$ is an
bilinear map $(X,Y) \mapsto \nabla_X Y$ associating to a pair of
vector fields $X,Y$ on $M$ another vector field $\nabla_XY$, satisfying:
\begin{enumerate}
\item $\nabla _{fX}Y=f\nabla _{X}Y$ for all functions $f$ on $M$;
\item $\nabla _{X}(fY)=df(X)Y+f\nabla _{X}Y$.
\end{enumerate}
Once this definition is given, it is possible to define $\nabla$
on tensors of every order.

On a Riemannian manifold we have a unique affine connection,
the {\sl Levi-Civita connection} 
$\nabla$, which is torsion free and preserves the metric, i.e. 
$\nabla g=0$.
In local coordinates the components of the connection are 
called the {\sl Christoffel symbols}:
$$
\nabla _{{\frac {\partial }
{\partial x^{i}}}}{\frac {\partial }
{\partial x^{j}}}=\Gamma ^{k}_{ij}{\frac {\partial }
{\partial x^{k}}}
$$
By the above  mentioned uniqueness, the $\Gamma ^{k}_{ij}$'s
are expressed in
terms of the metric components $g_{ij}$:
$$
{\displaystyle \Gamma _{jk}^{l}={\tfrac {1}{2}}g^{lr}\left(\partial _{k}g_{rj}+
\partial _{j}g_{rk}-\partial _{r}g_{jk}\right)}
$$
where as usual $g^{ij}$ are the coefficients of the dual metric tensor, i.e. 
the entries of the inverse of the matrix $(g_{kl})$. 
The torsion freeness is equivalent to the symmetry
$$
{\displaystyle \Gamma _{jk}^{l}=
\Gamma _{kj}^{l}.} 
$$

\medskip
In $\R^n$ the {\sl gradient} of a scalar 
function $f$ is the vector field characterized by 
the property: $\mathrm{grad}(f) \cdot v=D_v$ (we
shall denote it with $\nabla(f)$ whenever no confusion arises). In other words,
its scalar product (in the euclidean metric) with a tangent vector $v$
gives the directional derivative of $f$ along $v$. When $M$ is a Riemannian
manifold with metric $g$, the gradient of a function $f$ on $M$ is
defined in the same way, except that the scalar product is now given by $g$.
So, in local coordinates, we have:
$$
\nabla_g f={\frac {\partial f}{\partial x^{i}}}g^{ij}
\frac {\partial }{\partial x^{j}}
$$
Notice that when $g_{ij}=\de_{ij}$, we retrieve the usual definition
in $\R^n$.

We end our short summary of the key concepts, %we use in our exposition, with
with the notion of {\sl geodesic}.

\medskip
A geodesic $\gamma$ on a smooth manifold $M$ with an affine connection $\nabla$
is a curve defined by the following equation:
$$
{\displaystyle \nabla _{\dot {\gamma }}{\dot {\gamma }}=0} 
$$
Geometrically, this expresses the fact that the parallel transport, 
given by $\nabla$,
along the curve $\gamma$ preserves any tangent vector to the curve. 
In local coordinates this becomes:
$$
{\displaystyle {\frac {d^{2}\gamma ^{\lambda }}{dt^{2}}}+
\Gamma _{\mu \nu }^{\lambda }{\frac {d\gamma ^{\mu }}{dt}}
{\frac {d\gamma ^{\nu }}{dt}}=0\ ,}
$$
Notice that when the metric is constant, we have the familiar equation:
$$
{\frac {d^{2}\gamma ^{\lambda }}{dt^{2}}}=0
$$
that is, the geodesics are straight lines.

\end{document}